\documentclass[11pt, a4paper]{article}

\usepackage[utf8]{inputenc}   

\usepackage{amsmath}          
\usepackage{amsfonts}         
\usepackage{amssymb}          

\usepackage{graphicx}         
\usepackage{hyperref}         
\hypersetup{
    colorlinks=true,
    linkcolor=blue,
    filecolor=magenta,      
    urlcolor=cyan,
    citecolor=green,
}

\usepackage[margin=2.5cm]{geometry} 

\title{RAIL: An Automatic Classifier of the Artificial Intelligence Readiness Level}

\author{%
Juan Irving Vasquez\thanks{Corresponding author: \href{mailto:jvasquezg@ipn.mx}{jvasquezg@ipn.mx}}\\
\small CIETEC-IPN\\ \textit{Instituto Politécnico Nacional}\\ México City, México
\and
Juan Terven \\
\small CICATA-QRO\\ \textit{Instituto Politécnico Nacional}\\ Querétaro, México 
\and
Laura- Ivoone Garay-Jiménez,\\ \textit{Instituto Politécnico Nacional},\\ UPIITA,
México City, México
}
\date{August, 2026} 

\begin{document}

\maketitle 

\abstract{Assessing the maturity of artificial intelligence technologies is essential for investment decisions, project management, and policy monitoring, yet the available readiness frameworks are heterogeneous and difficult to apply automatically: the adaptation of Technology Readiness Levels to AI lacks AI-specific gating criteria, the Machine Learning Technology Readiness Levels presuppose access to internal process artifacts, and AI/data readiness dimension models employ scales that resist direct comparison. This paper makes two contributions. First, we unify these three frameworks into the Unified AI Readiness Level (AIRL), a nine-level ordinal scale built on an environmental evidence ladder and complemented by dimensional caps (covering specification, data existence, data quality, data legality, expert knowledge, and algorithmic maturity) together with a generality-anchoring rule and explicit assignment disciplines, so that a readiness level becomes decidable from a natural-language description of the work alone. Second, we propose RAIL (Readiness Assessment via Independent LLM-experts), a panel-of-experts classifier that operationalizes the scale: one evidence agent and six independent dimension agents, each a large language model with a narrowly scoped mandate, deliver verdicts that a deterministic minimum rule aggregates and a chief expert reviews under asymmetric authority, confirming or lowering the panel's recommendation but never raising it above the caps. The method was tested in the analysis of several research works showing consistency and avoiding overestimation from monolithic LLM classifiers. Code available at \url{https://github.com/irvingvasquez/RAIL}}

\section{Introduction}
\label{sec:introduction}

Artificial intelligence has moved from laboratory research to a strategic asset in industry, government, and science\cite{lecun2015deep}, and with this transition the question of \emph{maturity} has become as consequential as the question of performance. Investors must decide which prototypes are close enough to deployment to fund, project managers must plan the transition from research artifact to product, and policy bodies must monitor the state of national and institutional AI portfolios. All of these decisions presuppose the ability to place a given piece of work on a common maturity scale. For conventional engineered systems this role has long been played by the Technology Readiness Levels (TRL) introduced by NASA~\cite{sadin1989,mankins1995technology}, a
nine-level ordinal scale ranging from the observation of basic principles to a system proven in operation.

The TRL scale, however, was conceived for deterministic hardware and software, and a growing body of literature shows that it transfers poorly to artificial intelligence:
AI systems are data-dependent, stochastic, and sensitive to distribution shift, so that a system validated in the laboratory may degrade unpredictably once 
deployed~\cite{lavin2022technology, browne2024, browne2025}. In response, several AI-specific readiness frameworks have been proposed, from the
contextualization of the nine TRLs for AI by the European Commission's AI Watch initiative~\cite{martinez_plumed_2020_aiwatch}, through the Machine Learning Technology Readiness
Levels (MLTRL) that recast the progression as a gated systems-engineering process~\cite{lavin2022technology}, to dimensional models that decompose readiness into AI and data specific facets such as data quality and data
legality~\cite{eljasik2019}. While conceptually mature, these frameworks are heterogeneous and difficult to apply automatically: the AI adaptation of the TRL lacks
AI-specific gating criteria, MLTRL presupposes access to internal process artifacts that are rarely visible in a textual description of a work, and the dimensional models employ scales of differing lengths that resist direct comparison. In practice, therefore, readiness assignment remains a manual exercise performed by panels of subject-matter experts ~\cite{dastoor2023,jain2025,betancourt2025}.

Automatic estimation of readiness levels has been explored, but existing approaches fall short of the problem just described. Bibliometric and clustering methods estimate maturity only at the level of entire technology fields and at coarse granularity~\cite{dastoor2023,jain2025}; supervised ensembles depend on small, domain-specific labeled datasets~\cite{chukhray2022}; and 
recent single-LLM assistants, while promising, inherit the biases and hallucination risks of a single model and still deviate from expert judgment in a non-negligible fraction of cases~\cite{betancourt2025}. Moreover, virtually all of this work targets the legacy unidimensional TRL rather than the richer, multidimensional AI-readiness frameworks, whose dimensions each require specialized expertise. The literature thus exhibits a twofold gap: AI-specific frameworks lack automated, reproducible estimation mechanisms, and existing estimators fail to reproduce the the property that makes manual assessment reliable.

\begin{figure}
    \centering
    \includegraphics[width=0.8\linewidth]{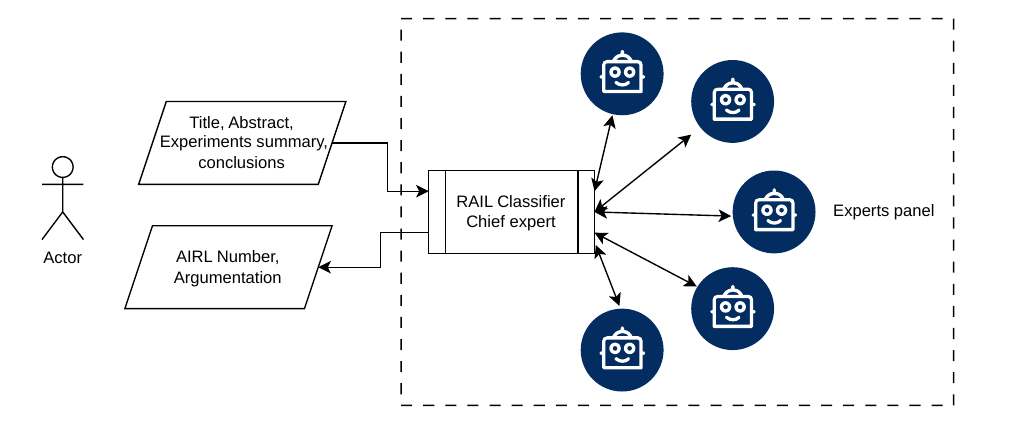}
    \caption{The RAIL classifier estimates a level of readiness for an AI application. It receives a natural language description of the work and estimates the level and provides an argumentation of the decisions. The decision is taken by a chief who valorates the comments of a panel of experts.}
    \label{fig:panel}
\end{figure}

This paper closes that gap with two contributions. First, we propose the \emph{Unified AI Readiness Level} (AIRL), an adapted nine-level ordinal scale that integrates
the EU interpretation of the TRL~\cite{martinez_plumed_2020_aiwatch}, the MLTRL of Lavin et al.~\cite{lavin2022technology}, and the dimensional model of Eljasik-Swoboda et al.~\cite{eljasik2019} into a single framework built on an environmental evidence ladder, complemented by dimensional caps over six AI-specific readiness
dimensions, a generality-anchoring rule, and explicit assignment disciplines, so that a readiness level becomes decidable from a natural-language description of the work alone. Second, we propose \emph{RAIL} (Readiness Assessment via Independent LLM-experts), a panel-of-experts classifier that operationalizes the scale (See Figure \ref{fig:panel}): one evidence expert and six independent dimension experts, each LLM with a narrowly scoped mandate, deliver verdicts that a deterministic minimum rule aggregates and a chief
expert reviews under asymmetric authority, confirming or lowering the panel's recommendation but never raising it above the caps. By construction, the resulting
classifier is sound with respect to the framework, conservative under ambiguity, neutral
under silence (lack of evidence), and auditable. We evaluate the approach on a corpus of graduate theses, comparing monolithic and panel-based classification under three readiness protocols, and show that the panel architecture avoids both the maturity inflation and
the missed dimensional gaps exhibited by monolithic LLM classifiers.

The remainder of the paper is organized as follows. Section~\ref{sec:related_work} reviews readiness frameworks and prior work on automatic readiness estimation. Section~\ref{sec:airl} presents the Unified AI Readiness Level and its assignment rules. Section~\ref{sec:rail} describes the RAIL panel architecture. Section~\ref{sec:experiments} reports the experimental evaluation, and Section~\ref{sec:conclusions} concludes.

\section{Related Work}
\label{sec:related_work}


The Technology Readiness Level (TRL) scale was introduced by NASA as a nine-level metric to assess the maturity of a technology, ranging from the observation of basic principles (TRL~1) to a system proven in an operational environment (TRL~9)~\cite{sadin1989,mankins1995}. Owing to its simplicity and domain independence, the scale was subsequently adopted by the European Commission, defense agencies, and innovation-management bodies as a standard instrument for acquisition and funding decisions. However, the original TRL was conceived for deterministic hardware and software systems, and a growing body of literature shows that it is ill-suited for artificial intelligence (AI): AI systems are data-dependent, stochastic, and sensitive to distribution shifts, so a system validated in a laboratory may degrade unpredictably once deployed~\cite{lavin2022,browne2024,browne2025}.

\subsection{From TRL to AI Readiness Level}
\label{subsec:trl_evolution}

Several adaptations of the TRL scale have therefore been proposed. Eljasik-Swoboda et al.~\cite{eljasik2019} extended the READINESSnavigator innovation-assessment tool with six AI- and data-specific readiness dimensions (e.g., algorithmic, data-quality, and data-legal readiness), replacing manufacturing-oriented criteria with data-centric ones. Martínez-Plumed et al.~\cite{martinezplumed2020}, within the European Commission's AI~Watch initiative, contextualized the nine TRLs for AI and introduced bidimensional \emph{readiness-versus-generality} charts, showing that high readiness is only attained by narrow, low-generality AI applications. Lavin and Renard~\cite{lavin2020} and later Lavin et al.~\cite{lavin2022} formalized the Machine Learning Technology Readiness Levels (MLTRL), a ten-level (0--9) systems-engineering framework featuring non-monotonic ``switchbacks,'' standardized TRL~Cards, and gated multidisciplinary reviews spanning research to continuous post-deployment monitoring.

More recent work has broadened readiness assessment beyond purely technical maturity. In the national-security domain, Browne et al.~\cite{browne2024} interviewed defense experts who deemed the classical TRL an ineffective metric for AI and proposed the AI Readiness Level (AIRL) framework, which gates TRL progression behind minimum thresholds in five dimensions: alignment, justified confidence, governance, human readiness, and data readiness; a companion study elaborates this multidimensional framework for military combat systems~\cite{browne2025}. Their case studies show that commercially mature systems (TRL~9) can score as low as AIRL~1 once AI-specific risks are considered. Müller et al.~\cite{muller2026} proposed the Use Case-Centered AI Readiness Level (UCAIRL), an eleven-level scale that makes domain-specific problem framing, data audits, and compliance mandatory prerequisites of technical implementation, addressing the paradox of technically mature AI projects that fail operationally. Domain- and organization-oriented adaptations follow the same trend: a unified nine-level TRL ladder for clinical AI derived through systematic review and Delphi synthesis~\cite{sayyari2025}, a TRL-based mapping of AI maturity in maternal health interventions~\cite{marquardt2025}, an organizational AI-readiness index for the public sector built on digital-transformation and data-management maturity~\cite{alfadhli2025}, and an enterprise-level AIRL scale grounded in diffusion-of-innovations and technology--organization--environment theory~\cite{garlatticosta2026}. Collectively, this evolution reflects a shift from a unidimensional, technology-centric scale toward multidimensional, socio-technical readiness models in which data, governance, human factors, and use-case context are first-class evaluation criteria.

\subsection{Automatic Estimation of Readiness Levels}
\label{subsec:automatic_estimation}

Regardless of the framework adopted, assigning a readiness level remains a predominantly manual exercise performed by panels of subject-matter experts, which is slow, costly, poorly scalable, and prone to subjectivity and inter-rater inconsistency~\cite{dastoor2023,jain2025,betancourt2025}. A second research stream has therefore explored the automatic estimation of readiness levels.

Early work framed the problem with soft-computing techniques---artificial neural networks, genetic algorithms, and fuzzy logic---to recognize readiness levels of R\&D projects under uncertainty~\cite{yusufova2022}. Chukhray et al.~\cite{chukhray2022} trained a stacking ensemble of weak regressors, with a random-forest meta-learner, on 56 university R\&D projects to estimate product readiness and commercialization cost, explicitly positioning the ensemble as a low-cost surrogate for a panel of human meta-experts. A complementary line of work exploits bibliometric signals as observable proxies of maturity: Dastoor et al.~\cite{dastoor2023} fitted S-curves to publications, patents, grants, and NASA Spinoff data for 31 technologies and applied ordinal regression to predict TRLs, while Jain and Kumar~\cite{jain2025} clustered 136 technology trends with unsupervised methods (MDBSCAN), classifying maturity into coarse \emph{watch}/\emph{prepare}/\emph{act} bands with 84.9\% accuracy and without labeled data. Most recently, Betancourt et al.~\cite{betancourt2025} leveraged large language models (LLMs), fine-tuning LLaMA~2 and GPT-3.5-Turbo on approximately 2{,}500 TRL-specific samples to build a conversational virtual assistant that estimates TRLs through natural-language dialogue; on eight real prototypes, the assistant matched expert assessments exactly in 50\% of cases and within one level in a further 37.5\%.

Despite this progress, existing automatic approaches exhibit important limitations. Bibliometric and clustering methods estimate maturity only at the level of entire technology fields, operate at a coarse granularity (often three bands rather than nine levels), and cannot assess an individual project from its own evidence~\cite{dastoor2023,jain2025}. Supervised ensembles depend on small, domain-specific labeled datasets that are expensive to curate and generalize poorly~\cite{chukhray2022}. Single-LLM assistants, while promising, inherit the biases and hallucination risks of a single model, rely on limited fine-tuning corpora, and still deviate from expert judgment in a non-negligible fraction of cases~\cite{betancourt2025}. Moreover, virtually all estimation work targets the classical TRL scale, and does not yet address the richer, multidimensional AI-readiness frameworks reviewed, whose dimensions (e.g., data, governance, and human readiness) each require specialized expertise.


The literature thus reveals a twofold gap. First, while AI-specific readiness frameworks have matured conceptually, they lack automated, scalable, and reproducible estimation mechanisms; conversely, existing automatic estimators remain tied to the legacy unidimensional TRL. Second, the very property that makes manual assessment reliable---the deliberation of a multidisciplinary panel of experts, as institutionalized in gated reviews~\cite{lavin2022} and Delphi syntheses~\cite{sayyari2025}---is precisely what current single-model estimators fail to reproduce. This motivates the model proposed in this work: a panel of LLM-based experts in which multiple specialized agents, each responsible for a distinct readiness dimension, independently assess the available evidence and aggregate their judgments. 

\section{The Unified AI Readiness Level}
\label{sec:airl}

The frameworks reviewed in the preceding section approach the maturity of artificial intelligence from several perspectives. However, we have restricted our research to three proposals. The adaptation of the NASA Technology Readiness Levels to AI proposed by Mart\'inez-Plumed et al.~\cite{martinez_plumed_2020_aiwatch} preserves the canonical nine-level environmental progression, from basic principles observed to an actual system proven in operation, and demonstrates that readiness in AI cannot be stated meaningfully without fixing the level of generality at which a technology is expected to perform. The Machine Learning Technology Readiness Levels of Lavin et al.~\cite{lavin2022technology} translate the same progression into an operational systems-engineering process for machine learning, contributing precise definitions of the intermediate transitions (proof of principle, proof of concept, capability, integration) as well as mechanisms, such as switchbacks and gated reviews, that acknowledge the non-monotonic character of real ML development. Finally, Eljasik-Swoboda et al.~\cite{eljasik2019} decompose readiness into AI-specific dimensions: specification, algorithmic maturity, and four facets of data readiness (existence, format and quality, legality, and expert knowledge). Then, they establish the principle that the overall readiness of an innovation is the minimum across its constituent dimensions. None of the three frameworks alone suffices for the classification task addressed in this work: the TRL scale lacks AI-specific gating criteria, ML-TRL presupposes access to internal process artifacts that are rarely visible in a textual description of a work, and the dimensional model of Eljasik-Swoboda et al.\ employs heterogeneous scales of five to nine levels that resist direct comparison. 

The Unified AI Readiness Level (AIRL) proposed here integrates the three into a single ordinal scale with explicit, evidence-based assignment rules, designed so that a human annotator or a large language model can assign a level to a natural-language description of an AI endeavour in a reproducible manner.

\subsection{Structure of the AIRL scale}
\label{sec:airl-structure}

The AIRL scale comprises nine ordinal levels whose primary discriminating variable is the \emph{evaluation environment} in which evidence of functioning has been produced. This choice follows directly from the environmental grouping implicit in the EU adaptation of the TRLs~\cite{martinez_plumed_2020_aiwatch}, where levels one through four correspond to laboratory conditions, five and six to relevant or simulated operational conditions, and seven through nine to the operational environment itself. The environment is preferred over alternative discriminators, such as reported predictive performance, because it is observable in textual descriptions, is largely domain-independent, and correlates with the residual risk that separates a demonstrated result from a dependable product: as both \cite{martinez_plumed_2020_aiwatch} and \cite{lavin2022technology} observe, benchmark performance does not directly translate into a technology that is ready for use in real-world environments.

At the lower end of the scale, AIRL~1 denotes work in which basic principles have been formulated but no experiment has been executed; it absorbs both TRL~1 \cite{mankins1995technology, iso162902013} and the theoretical ``first principles'' stage that MLTRL~\cite{eljasik2019} designates as Level~0. AIRL~2 requires that a concrete application concept exists and that exploratory experiments have been performed on sample, toy, or synthetic data, in correspondence with MLTRL's goal-oriented research stage, in which experiments probe specific model properties rather than end-to-end performance. AIRL~3 constitutes the experimental proof of principle: the approach is validated in a testbed, typically against public benchmarks or simulated data, with defined metrics and baseline comparisons; this level unifies TRL~3 with MLTRL Level~2 and marks the first point at which quantitative claims become admissible evidence. AIRL~4 requires that the constituent components (model, data pipeline, and interfaces) have been integrated and shown to work together in a controlled environment, with the codebase raised to what Lavin et al.\ term prototype caliber.

The middle of the scale captures the transition from research artifact to application. AIRL~5 is reached when the technology has been validated on real, representative data of the target use case in a relevant environment, and when the evaluation includes application-oriented measures in addition to conventional ML metrics; this level merges TRL~5 with the proof-of-concept stage of MLTRL, whose defining requirement is precisely the substitution of curated research data by noisy, real-world data. AIRL~6 corresponds to the demonstration of the technology as a \emph{capability}, in the sense given to that term by Lavin et al.: the model no longer operates in isolation but as a module of a larger workflow, demonstrated in a relevant environment before stakeholders beyond the research team, with product requirements and their verification and validation measures explicitly drafted. This level deliberately marks the research-to-product handoff that MLTRL identifies as the ``valley of death'' of applied machine learning \cite{balch2021bridging}, and that the READINESS navigator observations of Eljasik-Swoboda et al.\ associate with the structural bias of technically oriented organizations toward technology readiness at the expense of market-facing activities.

The upper levels concern operation. AIRL~7 requires an actual system prototype functioning in the operational environment, typically as a pilot, field trial, beta program, or shadow deployment on live data. AIRL~8 denotes the system complete and qualified: verified against the full set of requirements, subjected to deployment-oriented testing regimes such as canary or shadow tests, and, in regulated domains, certified by the competent authority (a condition that the EU TRL \cite{martinez_plumed_2020_aiwatch} rubric makes explicit and that we adopt as a hard requirement). AIRL~9, finally, is reserved for systems proven in sustained operation. Following Lavin et al., we require at this level not merely the fact of deployment but evidence of the maintenance apparatus that deployment of learning systems demands: monitoring for data drift, concept drift, and performance degradation, together with defined retraining or feedback processes. A system described as launched but lacking any monitoring provision is, under the unified scale, qualified rather than proven.

\subsection{Dimensional gating}
\label{sec:airl-gating}

The environmental ladder alone would misclassify a well-known failure mode of AI projects: work that exhibits advanced experimental evidence while resting on unresolved foundations, most commonly of a legal or data-related nature. To capture this, the AIRL incorporates the dimensional model of Eljasik-Swoboda et al.~\cite{eljasik2019} not as a set of parallel scales but as a system of \emph{caps} on the environmentally derived level. Six dimensions are monitored: the specification of the use case and its success criteria; the existence of and access to the required data; the understanding of data format and the measurement of data quality, including bias and class imbalance; the legality of data use, encompassing personal-data protection and freedom-to-operate considerations; the availability of captured expert knowledge where the task demands it; and the algorithmic dimension, covering the selection, evaluation on real data, and tuning of the learning method. When a description provides explicit evidence that one of these dimensions is deficient (for instance, that the legality of processing the underlying data remains unclear, or that a concrete learning target has not been defined) the assigned level is bounded above by the level compatible with that deficiency, irrespective of the sophistication of the reported experiments. Formally, the final level is the minimum of the environmentally evidenced level and the caps induced by explicitly evidenced dimensional gaps. This construction preserves the minimum principle of the source framework, in which an innovation cannot outrank its weakest readiness field, while adapting it to the epistemic situation of a classifier that observes only a textual description: silence with respect to a dimension is treated as neutral, and only affirmative evidence of a gap triggers a cap. The same minimum principle is applied compositionally, following the system-level rule of Lavin et al.\ whereby the readiness of a system equals the lowest readiness of its essential constituent parts; a pipeline that couples a mature, deployed component with a newly prototyped one is classified at the level of the latter.

\subsection{Generality anchoring and assignment discipline}
\label{sec:airl-rules}

A distinctive contribution of Mart\'inez-Plumed et al.~\cite{martinez_plumed_2020_aiwatch} is the observation that readiness and generality trade off against one another: a technology specialized to a narrow, controlled domain may attain the highest levels while its more general formulation remains at the stage of basic research, so that a readiness statement is ill-defined unless the scope of capability is fixed. The unified scale internalizes this observation as an anchoring rule rather than as a second axis. Each description is classified at the level of generality that it itself claims: when the demonstrated evidence pertains to a narrower scope than the claimed one, the level assigned is that which the evidence supports for the broad claim, which is in general lower; conversely, a description that explicitly restricts its own scope is evaluated within that restriction. This preserves the analytical content of the readiness-versus-generality charts while yielding the single ordinal label that a classification setting requires.

Three further disciplines govern assignment, all motivated by properties of the source frameworks. First, evidence takes precedence over intention: aspirational formulations contribute nothing to the level, which is determined solely by what has demonstrably been done, a direct consequence of the gated-review logic of MLTRL \cite{lavin2022technology}, in which graduation from a level requires satisfied criteria rather than declared plans. Second, ambiguity is resolved conservatively toward the lower of two adjacent candidate levels, reflecting the risk-averse posture that all three frameworks inherit from their systems-engineering lineage. Third, maturity is not inherited: the incorporation of a pretrained or off-the-shelf component does not transfer that component's readiness to the work under evaluation, and, symmetrically, a previously deployed system undergoing rework is classified at the current state of the reworked component, in accordance with the switchback mechanisms through which MLTRL \cite{lavin2022technology}formalizes regressions in maturity. Taken together, the environmental ladder, the dimensional caps, the generality anchoring, and the assignment disciplines define a classification function from textual descriptions to the ordinal set $\{1,\dots,9\}$ that is faithful to the three source frameworks while remaining decidable from the information a description actually contains.

\begin{table}[t]
\centering
\caption{Correspondence between the Unified AI Readiness Level (AIRL) and related frameworks. TRL levels follow the EU adaptation~\cite{martinez_plumed_2020_aiwatch}; MLTRL levels follow Lavin et al.~\cite{lavin2022technology}. The dimensions of Eljasik-Swoboda et al.~\cite{eljasik2019} act transversally as caps on all levels.}
\label{tab:airl-mapping}
\begin{tabular}{cp{0.7\textwidth}cc}
\hline
AIRL & Defining evidence & TRL & MLTRL \\
\hline
1 & Principles formulated; no experiment executed & 1 & 0 \\
2 & Concept with exploratory experiments on sample or synthetic data & 2 & 1 \\
3 & Proof of principle in testbeds or benchmarks with defined metrics & 3 & 2 \\
4 & Components integrated and validated in a controlled environment & 4 & 3 \\
5 & Validation on real, representative data in a relevant environment & 5 & 4 \\
6 & Capability demonstrated within a larger workflow; R\&D--product handoff & 6 & 5 \\
7 & System prototype operating in the operational environment (pilot) & 7 & 6--7 \\
8 & System complete, verified, and qualified or certified & 8 & 8 \\
9 & Sustained operation with monitoring and maintenance processes & 9 & 9 \\
\hline
\end{tabular}
\end{table}

\section{RAIL Classifier}
\label{sec:rail}

\begin{figure}
    \centering
    \includegraphics[width=\linewidth]{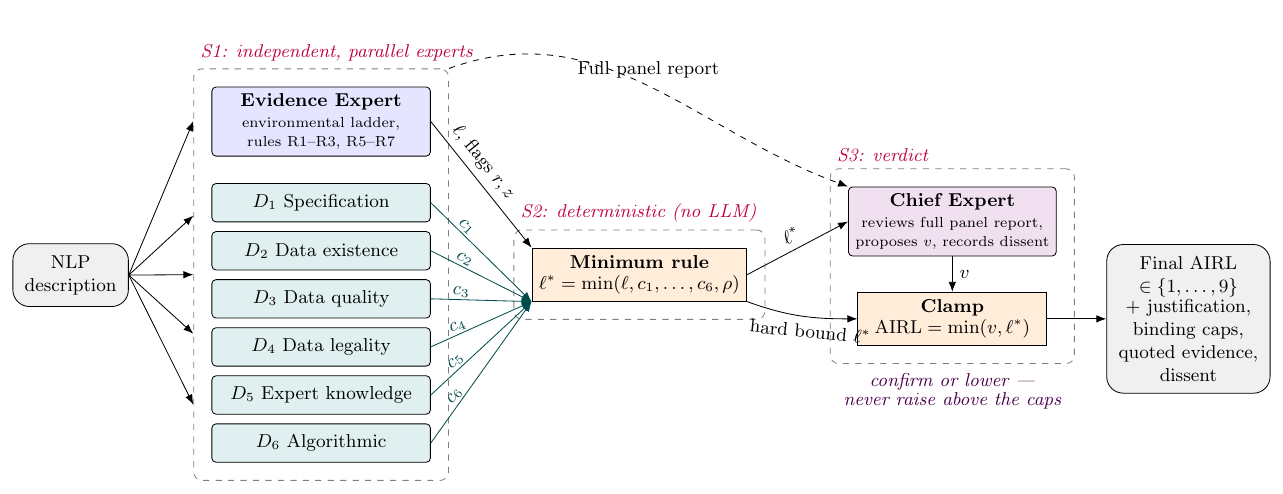}
    \caption{Workflow of the RAIL classifier. The description is evaluated by several experts. Then, a chief expert uses the of the experts an provides a final level.}
    \label{fig:placeholder}
\end{figure}

Assigning an AIRL to a natural-language description is, in principle, a task a single LLM could attempt in one pass \cite{ding2023parameter, sun2023text}: the rulebook of Section~\ref{sec:airl} could be supplied as an instruction and the model asked for an integer. In preliminary experimentation, however, such monolithic prompting conflates three judgments that the framework deliberately keeps distinct (the placement of the work on the environmental ladder, the detection of explicit gaps in the six readiness dimensions, and the application of the minimum rule that binds them) and it exhibits the two failure modes this conflation predicts. First, the model tends to \emph{hallucinate gaps}~\cite{huang2026generalization}, penalizing descriptions for dimensions about which they are merely silent, in violation of the neutrality-of-silence principle of Section~\ref{sec:airl-gating}. Second, and symmetrically, it tends to let strong experimental evidence \emph{argue past} an explicitly stated deficiency, allowing, for example, an impressive benchmark result to outweigh an unresolved data-legality question that the framework treats as a hard cap. Both failure modes stem from performing gating and evidence assessment within a single, holistic judgment. We therefore operationalize the classifier as a \emph{panel of experts}: a set of specialized agents, each instantiated as a LLM with a narrowly scoped mandate, whose individual verdicts are combined by a deterministic aggregation operator and reviewed by a presiding agent that issues the final verdict. The architecture mirrors the structure of the framework itself, so that each formal element of Section~\ref{sec:airl} is the explicit responsibility of exactly one component of the system.

\subsection{Panel composition}
\label{sec:panel-composition}

The panel comprises eight assessing agents and one deterministic operator, organized in three stages. In the first stage, a single \emph{evidence expert} and six \emph{dimension experts} examine the description independently and in parallel; none of them sees the output of any other. In the second stage, a deterministic aggregator, implemented in ordinary program logic rather than as a language model, combines their verdicts through the minimum rule. In the third stage, a \emph{chief expert} receives the complete panel report---every individual verdict together with the aggregated recommendation---and pronounces the final classification under an asymmetric authority constraint described below. The independence of the first stage is essential: because each dimension expert evaluates the description in isolation, a gap detected in one dimension cannot contaminate the judgment of another, and the evidence expert's placement on the ladder cannot be biased by knowledge of pending caps. This design corresponds to the epistemic situation the framework assumes, in which each readiness dimension is a property of the described work that either is or is not explicitly evidenced, independently of the others.

\subsection{The evidence expert}
\label{sec:evidence-expert}

The evidence expert is responsible solely for the environmental ladder of Section~\ref{sec:airl-structure}. Given the description, it identifies the most advanced evaluation environment for which demonstrated evidence exists (from paper studies through synthetic experimentation, benchmark testbeds, integrated laboratory pipelines, real-data validation, capability demonstration, operational piloting, and qualification, to monitored production) and returns a provisional level in $\{1,\dots,9\}$, together with the decisive evidence, quoted or closely paraphrased from the description. Because several of the assignment disciplines of Section~\ref{sec:airl-rules} require a holistic reading of the text rather than a dimensional one, they are assigned to this agent: the compositional minimum over essential components (R1), the exclusion of aspirational claims from the evidence base (R2), the non-inheritance of maturity from pretrained or off-the-shelf constituents (R5), the anchoring of the level to the claimed generality (R6), and the treatment of reworked, previously deployed systems at the current state of the reworked component (R7). Ambiguity between adjacent levels is resolved downward by instruction (R3). The evidence expert additionally emits two Boolean flags consumed later by the aggregator: whether the described application belongs to a regulated domain, and whether regulatory certification or approval is explicitly stated. It does not, by construction, evaluate any of the six readiness dimensions.

\subsection{The dimension experts}
\label{sec:dimension-experts}

Each of the six dimension experts is the specialist for exactly one dimension of Section~\ref{sec:airl-gating}: the specification of the use case and its success criteria ($D_1$), the existence of and access to the required data ($D_2$), the understanding of data format and the measurement of data quality including bias and imbalance ($D_3$), the legality of data use ($D_4$), the capture of required expert knowledge ($D_5$), and the algorithmic dimension covering selection, evaluation on real data, and tuning ($D_6$). A dimension expert does not output a level; it outputs a \emph{cap} $c_i \in \{9, 6, 4\}$, where $9$ denotes the absence of any cap, and $6$ and $4$ denote the two severities of explicitly evidenced deficiency defined by the framework. Two disciplines govern these agents and are enforced by both instruction and interface. The first is the neutrality of silence: an expert may impose a cap only upon affirmative textual evidence of a gap, and a description that does not mention the expert's dimension must receive $c_i = 9$; the criteria that qualify as affirmative evidence at each severity are enumerated verbatim in the agent's mandate, taken from the dimensional table of the framework. The second is evidentiary traceability: any cap must be accompanied by a verbatim quotation of the passage that triggered it, which is preserved in the classification record. At the interface level, the system adopts a fail-neutral policy: a malformed or unparsable expert response is interpreted as $c_i = 9$ rather than as a cap, so that infrastructure faults can never manifest as spurious readiness penalties.

\subsection{Deterministic aggregation}
\label{sec:aggregation}

The verdicts of the first stage are combined without recourse to a language model. Let $\ell \in \{1,\dots,9\}$ denote the provisional level returned by the evidence expert, $c_1,\dots,c_6$ the dimensional caps, $r \in \{0,1\}$ the regulated-domain flag, and $z \in \{0,1\}$ the certification flag. The aggregated recommendation is
\begin{equation}
\label{eq:minrule}
\ell^{\ast} \;=\; \min\Bigl(\, \ell,\; c_1,\; \dots,\; c_6,\; \rho \,\Bigr),
\qquad
\rho =
\begin{cases}
7 & \text{if } r = 1 \text{ and } z = 0,\\[2pt]
9 & \text{otherwise,}
\end{cases}
\end{equation}
where $\rho$ implements the regulated-domain discipline (R9), under which deployment-level language in a regulated domain cannot yield a level above seven in the absence of stated certification. Equation~\eqref{eq:minrule} is the direct formalization of the minimum principle inherited from Eljasik-Swoboda et al.~\cite{eljasik2019} and of the system-level composition rule of Lavin et al.~\cite{lavin2022technology}. Implementing it as ordinary program logic rather than as a prompted judgment yields three properties that a monolithic classifier cannot offer. The rule is \emph{exact}, in that the binding constraint is computed rather than estimated; it is \emph{auditable}, in that the output records which caps were binding, each with its quoted evidence; and it is \emph{immune to persuasion}, in that no rhetorical property of the input text can influence the arithmetic that combines the panel's findings.

\subsection{The chief expert and the final verdict}
\label{sec:chief-expert}

The final stage submits the complete deliberation to a presiding agent, the chief expert, which receives the description, the evidence expert's placement with its rationale and applied rules, every dimensional verdict with its evidence or its declared silence, and the aggregated recommendation $\ell^{\ast}$. The chief expert fulfils two functions. The first is quality control over the panel itself: reviewing whether the evidence expert credited aspirational claims, overlooked a compositional-minimum or generality-anchoring situation, or over-read weak evidence, and whether any dimensional cap was granted too leniently. The second is the production of the final justification: a concise statement, suitable as a dataset label rationale, citing the decisive evidence and any binding cap.

Crucially, the chief expert's authority is \emph{asymmetric}. Denoting its proposed level by $v$, the final classification is
\begin{equation}
\label{eq:clamp}
\mathrm{AIRL} \;=\; \min\bigl(v,\; \ell^{\ast}\bigr),
\end{equation}
so that the presiding agent may confirm the panel's recommendation or lower it, but may never raise it above the deterministic bound. The asymmetry is a soundness requirement rather than a heuristic. The dimensional caps of Equation~\eqref{eq:minrule} are hard constraints of the framework: a description whose data legality is explicitly unresolved is at most AIRL~4 regardless of any other merit, and permitting a final holistic judgment to override that bound would reintroduce, at the last stage, precisely the argue-past failure mode that the panel architecture was designed to eliminate. The constraint is therefore enforced twice, once as instruction within the chief expert's mandate and once as the clamp of Equation~\eqref{eq:clamp} in program logic, so that soundness does not depend on instruction-following alone. Disagreement is nonetheless informative: when the chief expert dissents---lowering the verdict, or objecting to a cap it must nevertheless respect---the dissent is recorded verbatim in the classification output. Over a corpus, the dissent record functions as a diagnostic of the panel, indicating which agents' mandates require refinement, in a manner analogous to the review switchbacks through which MLTRL feeds gate-level disagreement back into the development process~\cite{lavin2022technology}.

\subsection{Properties of the architecture}
\label{sec:properties}

The resulting classifier is a function from descriptions to $\{1,\dots,9\}$ with properties that follow from its construction rather than from empirical tuning. It is \emph{sound with respect to the framework}, in the sense that no output can exceed the bound implied by the evidenced environmental level and the explicitly evidenced dimensional gaps, by Equations~\eqref{eq:minrule} and~\eqref{eq:clamp}. It is \emph{conservative}, since ambiguity is resolved downward at the evidence stage and the final stage can only confirm or lower. It is \emph{neutral under silence}, because caps require affirmative quoted evidence and interface faults default to the absence of a cap. It is \emph{auditable}, since every classification carries the provisional level, the set of binding caps with their quotations, the applied assignment rules, and any dissent, which together constitute a complete justification trail; when the classifier is used to produce training data for a distilled single-model classifier, this trail supplies rationale annotations at no additional cost. Finally, the architecture is \emph{modular}: the first-stage agents are mutually independent and may be executed concurrently, individual mandates may be revised without retraining or re-prompting the remainder of the panel, and the deterministic core guarantees that such revisions alter only the judgments delegated to language models, never the logic that binds them.

\section{Experiments}
\label{sec:experiments}

For the evaluation of the proposed AIRL protocol and the RAIL classifier, we conducted a combinatorial experiment in which each configuration is defined by a readiness-level protocol and a classification strategy.

\begin{itemize}
    \item Protocols
    \begin{itemize}
        \item TRL. The original Technology Readiness Level (TRL) protocol proposed by Mankins~\cite{mankins1995technology}, encoded as a single prompt.
        \item AI TRL. The AI-specific adaptation proposed by Martinez-Plumed et al.~\cite{martinez_plumed_2020_aiwatch}, encoded as a single prompt.
        \item AIRL. The readiness-level protocol proposed in this work.
    \end{itemize}

    \item Classification strategies
    \begin{itemize}
        \item Monolithic LLM (Mono). A single large language model assigns the readiness level in a single inference using the corresponding protocol description as a prompt.
        \item RAIL. The proposed multi-agent architecture, in which a panel of expert agents collaboratively determines the readiness level. This strategy is only applicable to AIRL.
    \end{itemize}
\end{itemize}

The evaluation corpus consists of master's and doctoral theses developed at CIDETEC. Since no publicly available dataset pairs technical descriptions of AI systems with expert-assigned readiness levels, we constructed our own corpus from institutional documents for which we have the appropriate usage rights.

For each thesis, we extracted the title, abstract, experimental summary, and conclusions, as these sections contain the most relevant evidence for readiness assessment. The experimental sections were summarized using Gemini Pro to reduce their length while preserving the essential technical information.

All prompts, source data, and implementation details are available in our repository \url{https://github.com/irvingvasquez/RAIL}. Experiments were executed locally on a workstation equipped with an Intel Core i9 processor, an NVIDIA RTX 4090 GPU, and the Ollama implementation of Qwen3:32B.

Since no benchmark dataset with ground-truth readiness labels exists, conventional accuracy metrics cannot be computed. Instead, the results in Table~\ref{tab:placeholder} are analyzed from three complementary perspectives: (i) the agreement among protocols and classification strategies, (ii) the internal deliberation of the RAIL architecture through the justification traces generated by its expert panel, and (iii) the computational cost of the multi-agent approach.

The selected theses also provide face validity for the evaluation, as academic research projects typically conclude at the proof-of-concept or prototype stage rather than operational deployment. Consequently, readiness levels concentrated in the lower-to-middle range are consistent with the expected maturity of the evaluated works.

\begin{table}[tb]
    \centering
    \begin{tabular}{|l|p{1.4cm}|p{1.4cm}|p{1.4cm}|p{1.4cm}|l|}
    \hline
    Work & TRL~\cite{mankins1995technology} (Mono) & IA TRL~\cite{martinez_plumed_2020_aiwatch} (Mono) & AIRL (Mono) & 
    AIRL (RAIL) \\
    \hline
    Alvarez 2023    & 4 & 7 & 5 & 4  \\
    Gante 2023      & 4 & 5 & 3 & 3  \\
    Brito 2024      & 6 & 7 & 4 & 4  \\
    Vasquez 2009    & 5 & 7 & 5 & 5  \\
    Mendoza 2018    & 4 & 4 & 5 & 4  \\
    Vazquez 2019    & 5 & 6 & 5 & 5  \\
    Rodriguez 2019  & 6 & 7 & 5 & 5  \\
    Silva 2023      & 4 & 5 & 5 & 5  \\
    Olguin 2023     & 4 & 7 & 5 & 5  \\
    Lopez 2018      & 5 & 7 & 5 & 5  \\
    Average         & 4.7 & 6.2 & 4.7 & 4.5 \\
    \hline
    \end{tabular}
    \caption{Experimentation results. Several master and PhD thesis weree analyzed by automatic technoloy readiness classifiers. Mono: monolitic LLM.}
    \label{tab:placeholder}
\end{table}

\subsection{Agreement structure across configurations}
\label{sec:analysis-agreement}

The four configurations produce markedly different distributions over the same ten documents. The original TRL protocol under a monolithic classifier yields a mean of $4.7$ (range $4$--$6$), the AI-adapted TRL of Mart\'inez-Plumed et al.~\cite{martinez_plumed_2020_aiwatch} a mean of $6.2$ (range $4$--$7$, with seven as the modal value, assigned to six of the ten theses), the AIRL under a monolithic classifier a mean of $4.7$ (range $3$--$5$, with five assigned to eight documents), and the full RAIL configuration a mean of $4.5$ (range $3$--$5$). Three regularities stand out.

First, the AI-adapted TRL under monolithic prompting is systematically the most generous configuration: it exceeds the original TRL on nine of ten documents, by $+1.5$ levels on average, and it places the majority of the corpus at TRL~7, the operational-prototype level. The per-document justifications explain the mechanism. For the earliest thesis in the corpus (Vasquez 2009), the classifier assigns TRL~7 on the grounds that the algorithm reconstructed a single physical object with a real stereoscopic camera, while simultaneously acknowledging that the real-world test ``was limited to a single object, introducing uncertainty about broader generality.'' A single-instance laboratory demonstration is thereby promoted to the operational environment, and the generality reservation, although correctly \emph{noticed}, is expressed as commentary rather than as a level reduction. This is precisely the \emph{argue-past} failure mode anticipated in Section~\ref{sec:rail}: within a single holistic judgment, salient positive evidence (a real camera, a real object) outweighs a constraint (generality anchoring) that the prompt states but nothing enforces. That master's theses are not, as a population, one step from deployment suggests that these assignments overestimate maturity; the systematic $+1.7$-level gap between this configuration and RAIL quantifies the inflation.

Second, the AIRL protocol under a monolithic classifier exhibits the opposite pathology in attenuated form: compression toward the center of the scale. Eight of ten documents receive level~5, and the dimensional apparatus almost never fires. The monolithic classifier, asked to scan six dimensions and apply the minimum rule within one generation, tends to report all caps as neutral. The comparison with RAIL on the same protocol is instructive: the panel lowers two of the ten labels (from $5$ to $4$ in both cases) and equals the monolithic label on the remaining eight, so the two configurations agree within one level on the entire corpus, yet the disagreements are exactly the documents where an explicit dimensional gap exists in the text. For the tomato detection thesis (Alvarez 2023), the monolithic classifier assigns level~5 with no caps, whereas RAIL's independent $D_3$ expert locates and quotes the passage ``el conjunto de validaci\'on real carec\'ia de balance en sus clases'' (an explicitly stated, unmitigated class imbalance in the real validation set) and caps the label at~4. Decomposition thus increases the recall of explicitly stated gaps: the same model that overlooks the passage when judging everything at once retrieves it verbatim when its only mandate is data quality. See  It is worth noting that the quotation is returned in Spanish from an English-language prompt, indicating that the evidentiary-traceability mechanism operates across the corpus's bilingual composition.

\begin{figure}
    \centering
    \includegraphics[width=0.7\linewidth]{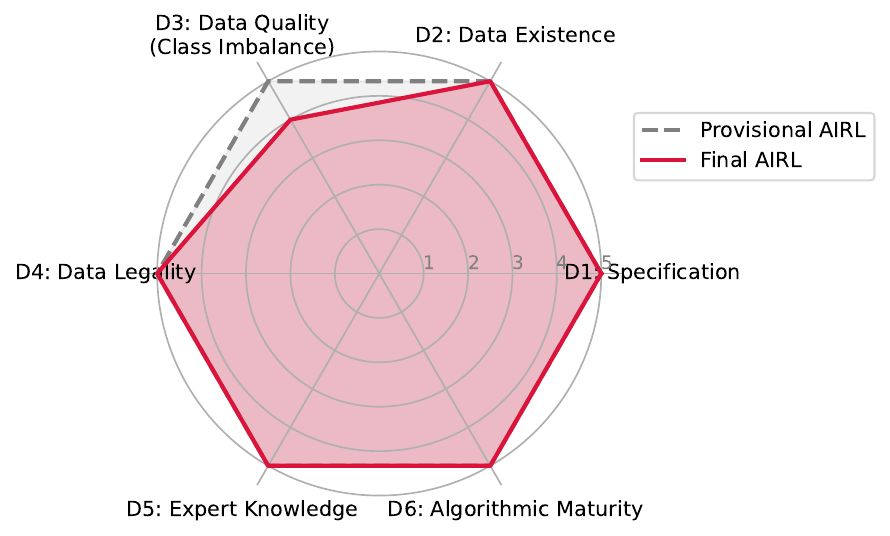}
    \caption{AIRL Multi-Agent Evaluation: The Alvarez thesis initially received a readiness score of 5. However, the Data Quality Expert identified unhandled class imbalance within the dataset. As a result, the final AIRL score was capped at 4.}
    \label{fig:placeholder}
\end{figure}

Third, RAIL is empirically conservative with respect to its monolithic counterpart: it never exceeds the monolithic AIRL label on any document. This is consistent with, though not implied by, the construction and it corroborates the design intent that the panel architecture removes upward pressure rather than adding downward noise.

\subsection{Panel-internal audit}
\label{sec:analysis-audit}

The per-agent records of the RAIL classifications permit an audit that no monolithic configuration supports, and the audit yields both confirmations and a finding of genuine methodological value.

On the confirmation side, the simulation-only thesis (Gante 2023) is the panel's cleanest case: the evidence expert places it at level~3, correctly registering R2 and R5, all six dimension experts return neutral caps, and the chief expert confirms with an empty dissent. The cloud-deployed segmentation thesis (Brito 2024) exercises the opposite path: the evidence expert reads the public deployment of the model on a serverless platform as an operational prototype and proposes level~7 (itself a debatable over-read, since a publicly reachable inference endpoint is not a pilot with real users) but the deterministic layer caps the label at~4, and the chief expert's dissent field records exactly the tension the architecture is designed to surface: the deployment suggests operational proximity, yet the binding cap and the generality anchoring override it. The dissent record thus functions as intended, preserving a disagreement that a single scalar label would erase. Across the audited classifications the chief expert confirmed the deterministic recommendation in every case and never lowered it, so the over-eager-chief failure mode did not materialize in this sample.

\subsection{Computational cost}
\label{sec:analysis-cost}

Decomposed deliberation multiplies inference. On the experimental platform (a single RTX~4090 running Qwen3-32B through Ollama, sequential execution, fixed seed), the monolithic baseline classifies a document in approximately $22$ seconds at a mean of roughly $4{,}000$ prompt-plus-completion tokens, while a full RAIL classification requires approximately $131$ seconds per document, a factor of six. For the intended use of the classifier, the production of audited training labels in batch, the cost is incurred once per document and buys the justification trail analyzed above; for interactive use, the monolithic AIRL configuration with the deterministic post-check of Section~\ref{sec:analysis-audit} offers a defensible low-cost operating point, at the price of the reduced cap recall documented in Section~\ref{sec:analysis-agreement}.

\subsection{Threats to validity}
\label{sec:analysis-threats}

The principal limitations of this evaluation are its scale and the absence of ground truth. Ten documents from a single institution, in a single genre (graduate theses in robotics and applied computer vision, in Spanish and English), classified by a single model at a single seed, support an analysis of agreement structure and mechanism, not a claim of accuracy; the systematic inflation of the AI-adapted TRL baseline, for example, is established relative to the face validity of the corpus rather than against expert labels. The summarization of the experimental sections by a separate large language model introduces a further mediation between the theses and the classifiers, common to all configurations but unquantified.

\section{Conclusions}
\label{sec:conclusions}

In this study we have addressed the problem of assessing the maturity of artificial intelligence
technologies from natural-language descriptions. We
introduced the Unified AI Readiness Level (AIRL), a nine-level ordinal scale that integrates previous protocols. Second, we proposed RAIL, a panel-of-experts classifier that operationalizes the scale.

The experimental evaluation over a corpus of graduate theses confirmed the failure modes that motivated the architecture. A monolithic classifier prompted with an AI-adapted TRL rubric systematically inflated maturity, exceeding the original TRL baseline; a monolithic classifier prompted with the full AIRL rulebook compressed its outputs toward the center of the scale and almost never activated the dimensional caps. In contrast, the proposed RAIL never exceeded its monolithic counterpart, and its independent dimension experts
recovered explicitly stated gap that the same underlying model overlooked when judging holistically. The dissent mechanism preserved informative disagreements between
the chief expert and the deterministic bound, providing a diagnostic signal that a single scalar label would erase. These benefits come at a computational cost of roughly six
times that of a single inference, a price that is justified for the batch production of audited training labels, while the monolithic AIRL configuration with a deterministic post-check remains a defensible low-cost alternative for interactive use.

The evaluation also delimits the scope of our claims. The corpus comprises ten documents from a single institution and genre, classified by a single model at a single seed, and no expert-assigned ground truth is available; consequently, our results establish agreement structure and mechanism rather than accuracy. Future work will proceed along three lines: the construction of a larger, multi-institutional corpus with expert-labeled readiness levels to enable proper accuracy and inter-rater agreement studies. 

\bibliographystyle{plain}
\bibliography{referencias}

@techreport{martinez_plumed_2020_aiwatch,
  title     = {AI Watch: Assessing Technology Readiness Levels for Artificial Intelligence},
  author    = {Martínez-Plumed, Fernando and Gómez-Gutiérrez, Emilio and Hernández-Orallo, José},
  year      = {2020},
  type      = {EUR},
  number    = {30401 EN},
  institution = {Publications Office of the European Union},
  address   = {Luxembourg},
  isbn      = {978-92-76-22987-2},
  doi       = {10.2760/15025},
  url       = {https://publications.jrc.ec.europa.eu/repository/handle/JRC122014}
}

@article{lecun2015deep,
  title={Deep learning},
  author={LeCun, Yann and Bengio, Yoshua and Hinton, Geoffrey},
  journal={nature},
  volume={521},
  number={7553},
  pages={436--444},
  year={2015},
  publisher={Nature Publishing Group UK London}
}

@article{lavin2022technology,
  title={Technology readiness levels for machine learning systems},
  author={Lavin, Alexander and Gilligan-Lee, Ciar{\'a}n M and Visnjic, Alessya and Ganju, Siddha and Newman, Dava and Ganguly, Sujoy and Lange, Danny and Baydin, At{\'\i}l{\'\i}m G{\"u}ne{\c{s}} and Sharma, Amit and Gibson, Adam and others},
  journal={Nature Communications},
  volume={13},
  number={1},
  pages={6039},
  year={2022},
  publisher={Nature Publishing Group UK London}
}

@misc{mankins1995technology,
  title={Technology readiness levels},
  author={Mankins, John C and others},
  year={1995},
  publisher={Nasa Washington, DC, USA}
}

@misc{iso162902013,
  author = {International Standar Organization},
  title       = {Space systems — Definition of the Technology Readiness Levels (TRLs) and their criteria of assessment},
  organization= {International Organization for Standardization},
  number      = {ISO 16290:2013},
  year        = {2013},
  address     = {Geneva, Switzerland},
  url         = {https://www.iso.org/standard/56064.html}
}

@article{balch2021bridging,
  title={Bridging the artificial intelligence valley of death in surgical decision-making},
  author={Balch, Jeremy and Upchurch, Gilbert R and Bihorac, Azra and Loftus, Tyler J},
  journal={Surgery},
  volume={169},
  number={4},
  pages={746--748},
  year={2021},
  publisher={Elsevier}
}

@article{ding2023parameter,
  title={Parameter-efficient fine-tuning of large-scale pre-trained language models},
  author={Ding, Ning and Qin, Yujia and Yang, Guang and Wei, Fuchao and Yang, Zonghan and Su, Yusheng and Hu, Shengding and Chen, Yulin and Chan, Chi-Min and Chen, Weize and others},
  journal={Nature machine intelligence},
  volume={5},
  number={3},
  pages={220--235},
  year={2023},
  publisher={Nature Publishing Group UK London}
}

@inproceedings{sun2023text,
  title={Text classification via large language models},
  author={Sun, Xiaofei and Li, Xiaoya and Li, Jiwei and Wu, Fei and Guo, Shangwei and Zhang, Tianwei and Wang, Guoyin},
  booktitle={Findings of the Association for Computational Linguistics: EMNLP 2023},
  pages={8990--9005},
  year={2023}
}

@article{huang2026generalization,
  title={Generalization or hallucination? understanding out-of-context reasoning in transformers},
  author={Huang, Yixiao and Zhu, Hanlin and Guo, Tianyu and Jiao, Jiantao and Sojoudi, Somayeh and Jordan, Michael and Russell, Stuart J and Mei, Song},
  journal={Advances in Neural Information Processing Systems},
  volume={38},
  pages={139807--139855},
  year={2026}
}

@techreport{sadin1989,
   author = {Sadin, Stanley R. and Povinelli, Frederick P. and Rosen, Robert},
   title  = {The {NASA} technology push towards future space mission systems},
   journal= {Acta Astronautica},
   volume = {20},
   institution = {NASA},
   pages  = {73--77},
   year   = {1989}
 }

@techreport{mankins1995,
   author      = {Mankins, John C.},
   title       = {Technology Readiness Levels: A White Paper},
   institution = {NASA, Office of Space Access and Technology},
   year        = {1995}
 }

@inproceedings{eljasik2019,
   author    = {Eljasik-Swoboda, Tobias and Rathgeber, Christian and Hasenauer, Rainer},
   title     = {Assessing Technology Readiness for Artificial Intelligence and Machine Learning based Innovations},
   booktitle = {Proceedings of the 8th International Conference on Data Science, Technology and Applications (DATA)},
   pages     = {281--288},
   year      = {2019}
 }

@techreport{martinezplumed2020,
   author      = {Mart{\'i}nez-Plumed, Fernando and G{\'o}mez, Emilia and Hern{\'a}ndez-Orallo, Jos{\'e}},
   title       = {{AI Watch}: Assessing Technology Readiness Levels for Artificial Intelligence},
   institution = {Publications Office of the European Union, Joint Research Centre},
   year        = {2020}
 }

@article{lavin2020,
   author  = {Lavin, Alexander and Renard, Gregory},
   title   = {Technology Readiness Levels for {AI} \& {ML}},
   journal = {arXiv preprint arXiv:2006.12497},
   year    = {2020}
 }

@article{lavin2022,
   author  = {Lavin, Alexander and Gilligan-Lee, Ciar{\'a}n M. and Visnjic, Alessya and Ganju, Siddha and Newman, Dava and Ganguly, Sujoy and Lange, Danny and Baydin, At{\i}l{\i}m G{\"u}ne{\c{s}} and Sharma, Amit and Gibson, Adam and Zheng, Stephan and Xing, Eric P. and Mattmann, Chris and Parr, James and Gal, Yarin},
   title   = {Technology readiness levels for machine learning systems},
   journal = {Nature Communications},
   volume  = {13},
   number  = {1},
   pages   = {6039},
   year    = {2022}
 }

@article{browne2024,
   author  = {Browne, Samuel T. and Pike, Thomas D. and Bailey, Mark M.},
   title   = {A Proposed Framework for Artificial Intelligence Safety and Technology Readiness Assessments for National Security Applications},
   journal = {OSF Preprints},
   year    = {2024}
 }

@article{browne2025,
   author  = {Browne, S. Tucker and Bailey, Mark M.},
   title   = {Rethinking Technological Readiness in the Era of {AI} Uncertainty},
   journal = {arXiv preprint arXiv:2506.11001},
   year    = {2025}
 }

@article{alfadhli2025,
   author  = {Alfadhli, Muna and Onat, Nuri C. and Kucukvar, Murat and Al-Maadeed, Somaya},
   title   = {Analyzing {AI} Readiness through Digital Transformation and Data Management: A Case Study of {Qatar}'s Government Sector},
   journal = {Applied Mathematics \& Information Sciences},
   volume  = {19},
   number  = {3},
   pages   = {497--507},
   year    = {2025}
 }

@article{sayyari2025,
   author  = {Sayyari, Mahsa and Karimi, Hasti and Baghi Keshtan, Sina and Saleknezhad, Atousa and Bemana, Reza and Rezaie, Iraj},
   title   = {Toward a Unified Technology Readiness Ladder for Clinical Artificial Intelligence: A Systematic Review and {Delphi} Synthesis},
   journal = {InfoScience Trends},
   year    = {2025}
 }

@article{marquardt2025,
   title={From pilot to practice: a scoping review protocol mapping the development of AI-enabled solutions for maternal health using technology readiness levels},
   author={Marquardt, Nico and Choi, Vladimir and Martyn-Dickens, Charles and Gorgens, Marelize and Mathewlynn, Sam and Kurth, Tobias and Bouteiller, Philipp and Wieler, Lothar H},
   journal={BMJ open},
   volume={15},
   number={8},
   pages={e105622},
   year={2025},
   publisher={British Medical Journal Publishing Group}
}

@inproceedings{muller2026,
   author    = {M{\"u}ller, Benedikt and Roth, Daniel and Kreimeyer, Matthias},
   title     = {Reframing {AI} readiness: a multi-dimensional use case-centered {AI} readiness framework},
   booktitle = {Proceedings of the Design Society, Volume 6: DESIGN 2026},
   year      = {2026}
 }

@article{garlatticosta2026,
   author  = {Garlatti Costa, Grazia and Pugliese, Roberto and Venier, Francesco},
   title   = {Exploring artificial intelligence adoption among {Italian} firms: the {AI} readiness level},
   journal = {International Journal of Business Information Systems},
   volume  = {51},
   number  = {7},
   year    = {2026}
 }

@inproceedings{yusufova2022,
   author    = {Yusufova, O. M. and Nevredinov, A. R.},
   title     = {An approach to applying soft computing models to recognize technology readiness levels in research and development ({R\&D}) projects},
   booktitle = {AIP Conference Proceedings},
   volume    = {2383},
   pages     = {070006},
   year      = {2022}
 }

@article{chukhray2022,
   author  = {Chukhray, Nataliya and Shakhovska, Nataliya and Mrykhina, Oleksandra and Lisovska, Lidiya and Izonin, Ivan},
   title   = {Stacking Machine Learning Model for the Assessment of {R\&D} Product's Readiness and Method for Its Cost Estimation},
   journal = {Mathematics},
   volume  = {10},
   number  = {9},
   year    = {2022}
 }

@inproceedings{dastoor2023,
author = {Jehan Dastoor and Heying Zhang and Michael G. Balchanos and Dimitri N. Mavris},
title = {A Bibliometric Approach to Characterizing Technology Readiness Levels Using Machine Learning},
booktitle = {AIAA SCITECH 2023 Forum},
chapter = {},
pages = {},
year ={2023},
doi = {10.2514/6.2023-2686},
URL = {https://arc.aiaa.org/doi/abs/10.2514/6.2023-2686},
eprint = {https://arc.aiaa.org/doi/pdf/10.2514/6.2023-2686}
}

@inproceedings{jain2025,
   author    = {Jain, Bhavesh Mahender and Kumar, Deepak},
   title     = {Predicting Tech Readiness through Bibliometric Analysis using Unsupervised Machine Learning},
   booktitle = {Proceedings of the XXXVI ISPIM Innovation Conference},
   year      = {2025}
 }

@article{betancourt2025,
   author  = {Betancourt, Juan and Coral, Andr{\'e}s and Fraga, Anabel and Figueroa, Cristhian and Ramirez-Gonzalez, Gustavo},
   title   = {Intelligent Virtual Assistant for Calculating Technology Readiness Levels Using Large Language Models ({LLM})},
   journal = {IEEE Access},
   volume  = {13},
   year    = {2025}
 }


\end{document}